\documentclass[runningheads]{llncs}
\pdfoutput=1
\usepackage{graphicx}
\usepackage{url}
\usepackage{color}
\usepackage{hyperref}
\usepackage[table,xcdraw]{xcolor}
\usepackage{amsmath}
\DeclareMathOperator{\sign}{sign}
\DeclareMathOperator{\argmin}{argmin}
\DeclareMathOperator{\argmax}{argmax}
\newcommand{\uargmin}[1]{\underset{#1}{\argmin}\;}
\newcommand{\uargmax}[1]{\underset{#1}{\argmax}\;}
\usepackage{amsfonts}
\usepackage{caption}
\usepackage{graphicx}
\usepackage{booktabs}

\usepackage{floatrow}
\newfloatcommand{capbtabbox}{table}[][\FBwidth]

\begin{document}
\title{GraphX$^\textbf{\small NET} -$ Chest X-Ray Classification Under Extreme Minimal Supervision}
%


%
\author{
Angelica I. Aviles-Rivero\inst{*1} \and
Nicolas Papadakis \thanks{Equal Contribution}\inst{2}   \and
Ruoteng Li \inst{3} \and
Philip Sellars \inst{1} \and
Qingnan Fan \inst{4} \and
Robby T Tan\inst{3,5} \and
Carola-Bibiane Sch\"onlieb\inst{1}
}
\authorrunning{A.I Aviles-Rivero et al.}

\institute{DPMMS and DAMPT, Faculty of Mathematics, University of Cambridge, UK  \email{\{ai323,ps644,cbs31\}@cam.ac.uk}\and
IMB, Universite de Bordeaux \email{nicolas.papadakis@math.u-bordeaux.fr} \and
National University of Singapore, Singapore \email{liruoteng@u.nus.edu} \and
Stanford University, USA \email{fqnchina@gmail.com} \and
Yale-NUS College, Singapore \email{robby.tan@yale-nus.edu.sg}
}



\maketitle

\begin{abstract}
The task of classifying X-ray data is a problem of both theoretical and clinical interest. Whilst supervised deep learning methods rely upon huge amounts of labelled data, the critical problem of achieving a good classification accuracy when an extremely small amount of labelled data is available has yet to be tackled. In this work, we introduce a novel semi-supervised framework for X-ray classification which is based on a graph-based optimisation model. To the best of our knowledge, this is the first method that exploits graph-based semi-supervised learning for X-ray data classification. Furthermore, we introduce a new multi-class classification functional with carefully selected class priors which allows for a smooth solution that strengthens the synergy between the limited number of labels and the huge amount of unlabelled data. We demonstrate, through a set of numerical and visual experiments, that our method produces highly competitive results on the ChestX-ray14 data set whilst drastically reducing the need for annotated data.

\end{abstract}

\section{Introduction}
The Chest X-Ray (CXR) is the most commonly performed x-ray examination which captures details of the lungs, heart, bones and blood vessels. CXRs play a critical role in diagnosing and monitoring conditions such as pneumonia, heart problems and lung cancer. However, it remains one of the most complex imaging studies to interpret~\cite{folio2012chest}. The effectiveness and accuracy of the interpretation heavily relies on the radiologist's expertise and still there is a substantial clinical error on the outcome~\cite{bruno2015understanding}. Furthermore, the requirement of human expertise increases the finical cost and time required for evaluation. Therefore, there is a clear need for fast automated evaluations of CXRs.

CXR classification has been widely addressed by the community, yet it remains an open problem. Early developments were based in handcrafted classification e.g.~\cite{toriwaki1973pattern}. However, this set of algorithmic approaches require particular modelling hypothesis to be met (e.g. texture, geometry, intensity), which may not be feasible to fulfill in practice. Due to the incredible results produced by deep learning in the field of computer vision, there has been a rush to apply deep learning architectures to the classification of CXRs \cite{yao2018weakly,wang2017chestx,bar2015chest}, which have shown promising results. The majority of these methods utilise deep convolutional neural network with architectures
such as ResNet \cite{resnet2015}, due to the success of these architectures in computer vision classification tasks. Several training methods have been considered including: pre-trained networks, fine tuned networks and networks trained from scratch on X-ray data e.g.~\cite{yao2018weakly,wang2017chestx,bar2015chest}.

However, a major drawback of these techniques is the high dependence on a large corpus of labelled data. Particularly in the medical domain, this might be a strong assumption for a solution, as annotated data contains strong human bias. Although there has been a huge effort in the community to mitigate this drawback by providing datasets such as ChestX-ray14, the has annotations but is far from being a definite expression of ground truth \cite{kohli2017medical}.  Therefore, by using supervised learning techniques one allows the labelling error and uncertainty to adversely effect the classification output of our machine learn framework.
To tackle both the effect of human bias and the limited amount of labelled data, we propose using the power of semi-supervised learning and graph representations.

\textbf{Our Contributions.} We propose a novel semi-supervised graph-based framework called GraphX$^{\textbf{\scriptsize NET}}$. Our contributions are: 1) \textit{a new multi-class classification functional with carefully chosen class priors}. Our framework is based on the normalised and non-smooth $p=1$ Laplacian.
2) We demonstrate that our novel framework learns to accurately classify CXRs, with a performance comparable to state-of-the-art deep learning techniques, whilst using an extremely smaller amount of labelled data.  3) This work also represents the first time that graph representations have been used for X-ray classification.

\begin{figure}[t!]
\centering
\includegraphics[width=1\textwidth]{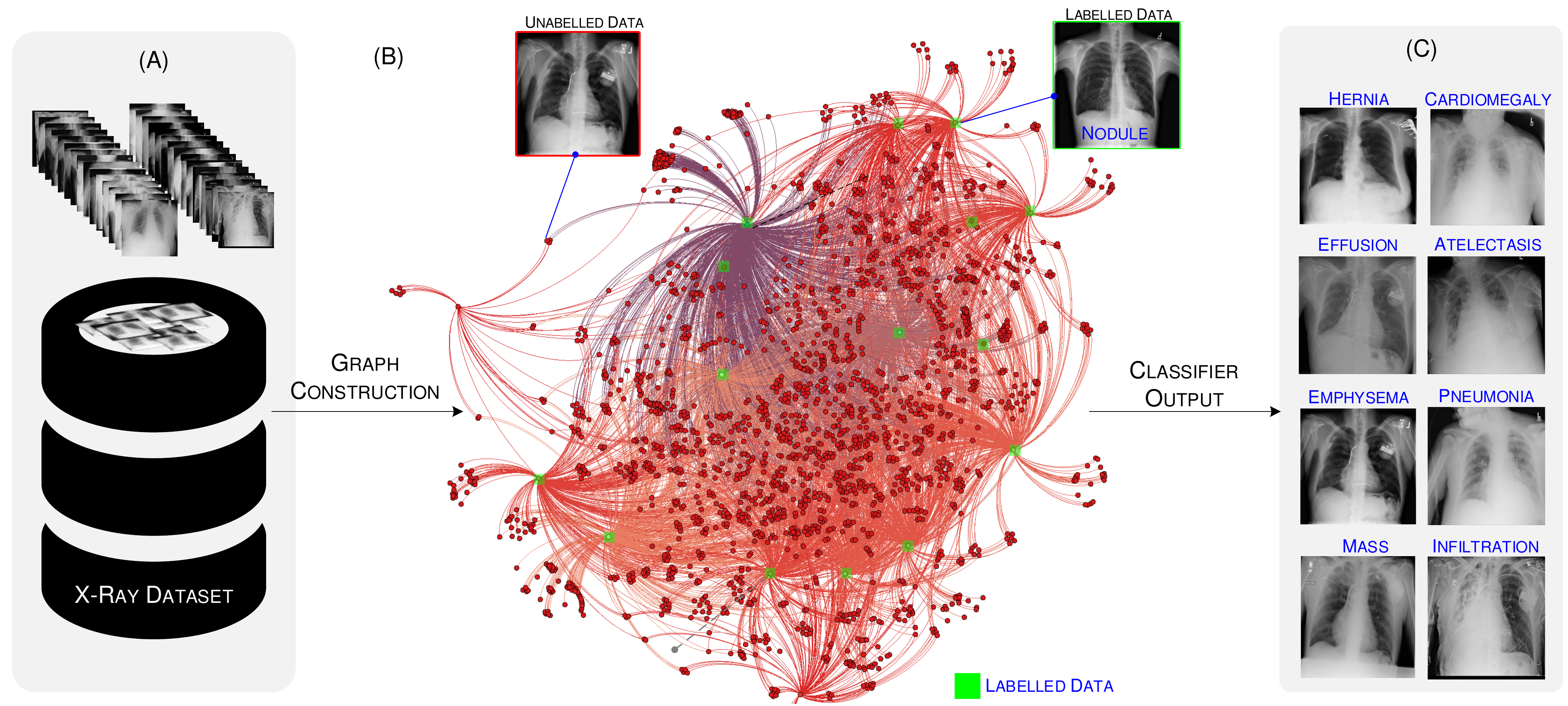}
\caption{Overview of our proposed GraphX$^{\textbf{\scriptsize NET}}$ method. We exploit both labeled and unlabeled data to produce high classification accuracy. In this framework, we aim to propagate labels for the unlabeled data with minimal supervision.}
\label{fig::teaser}
\end{figure}

\section{GraphX$^{\textbf{\scriptsize NET}}$ Framework for X-Ray Data Classification.}

Our approach is motivated by a central problem in medical imaging which is the lack of reliable quality annotated data. Although, transfer learning~\cite{bar2015chest} or Generative Adversarial Networks~\cite{moradi2015machine} somewhat mitigate this problem, they fail to account for the mismatch between expert annotation and ground truth annotation created by human bias and uncertainty. With this motivation in mind, we propose, for the first time, using a semi-supervised framework, call GraphX$^{\textbf{\scriptsize NET}}$ (see  Fig.~\ref{fig::teaser} for illustration).

\textbf{Data Representation with Graphs.} Although there are different methods for representing  data including conventional grid form. In this work, we motivate the use of graph data representations as follows. Firstly, graphs are a natural representation for groups of images where each node represents an individual image. Secondly, given that graph based methods seek to find smooth solutions to the created embedding, they are able to correct for initially mislabelled samples. Lastly, graph has strong and mathematical properties such as sparseness which allows for fast computation.

We represent a given dataset as an undirected weighted graph $\mathcal{G}=(\mathcal{V},\mathcal{E},W)$  compromising a set of $n$ nodes $\mathcal{V}$ which are connected by a set of edges $\mathcal{E}$ with weights $w_{ij}=S(i,j)\geq 0$ that correspond to some similarity measure $S$ between the features of nodes $i \in \mathcal{V}$ and $j \in \mathcal{V}$, $w_{ij}=0$  if  $(i,j)  \notin  \mathcal{E}$; and functions $u\in
\mathbb{R}^{n}$. Our setting is based on the normalised graph p-Laplacian, which reads:
\begin{equation}\label{model}
\Delta_p (u)=\sum_{i,j} w_{ij}\left\| \frac{u_i}{d_i^{1/p}}-\frac{u_j}{d_j^{1/p}}\right\|^p, \vspace{0.5cm} \textnormal{ with } p\geq 1 \textnormal{ and } d_i=\sum_j w_{ij}>0, \vspace{-0.5cm}
\end{equation}
where $d_i$ is the degree of node $i$. The eigenfunctions of the graph Laplacian operator give interesting understanding of the substructures of the graph.  Eigenfunctions of a normalised graph Laplacian for $p=2$ have been successfully used in different applications such as in~\cite{belkin2003Laplacian,chen2013inferring,dodero2014group}.

\smallskip

\textbf{Learning to Classify under Extreme Minimal Supervision.} However, unlike those works, our framework has a different aim which is to solely obtain classification estimates on the unlabeled samples. That is, to perform a node classification task on $\mathcal{G}$ with $L$ available classes, given an extremely small amount of labelled nodes $x_i$. More precisely,  given a small amount of labeled data $\{ (x_i ,y_i) \}_{i=1}^{l}$ with provided labels $\mathcal{L} = \{1,..,L\}$  and $\{y_i\}_{i=1}^{l} \in \mathcal{L}$ and a large amount of unlabelled data  $\{ x_k \}_{k=l+1}^{n}$, we seek to infer a function $f: \mathcal{X}^{n} \mapsto \mathcal{Y}^{n}$ such that $f$ gets a good estimate for  $\{ x_k \}_{k=l+1}^{n}$.

Although several works have explored this learning style, either from a pure machine learning perspective e.g.\cite{zhu2003semi} or a medical imaging perspective e.g. \cite{wang2016progressive}, these methods seek to only approximate $p\to 1$ in the graph Laplacian. However, recent developments on machine learning showed that the use of the unnormalised (i.e. without the re-scaling by the node degrees in  \eqref{model}) and non smooth $p=1$ Laplacian, related to total variation, can achieve better performance \cite{Buhler2009}.

To mitigate these current drawbacks in the literature, we propose a novel semi-supervised framework, GraphX$^\textbf{\small NET}$, based on the normalised and non smooth $p=1$ Laplacian in \eqref{model}. The function can then be rewritten as: $\Delta_1(u)=|WD^{-1}u|,$ where $W$ is the weight matrix $w_{ij}$ and $D$ the diagonal matrix containing the degrees $d_i$. To this end, we generalise the unsupervised binary normalised graph method of~\cite{Feld2019} to a semi-supervised multi-class graph approach. To this aim, our algorithmic approach is as follows.

For each class, $k=1\cdots L$, we consider a variable $u^k$ that has values for all nodes of the graph. For all unlabeled nodes $i>l$, the $L$ variables are then coupled with the constraints that for all nodes $i$:
$\sum_{k=1}^L u^k_i=0,\, \forall i>l.$
This simple coupling indeed leads to faster projection algorithms than simplex  \cite{bresson2013multiclass,gao2015medical} or non convex orthogonality constraints between $u^k$'s \cite{dodero2014group}. We assume that a set of annotated nodes $\mathcal I_k\subset\{1\cdots l\}$ are available for each class $k$: $y_i=k\in\mathcal{L}$ for all $i\in\mathcal{I}_k$. Taking a small parameter $\epsilon>0$, we therefore constrain that:
\begin{equation}\label{constr_data}
\left\{\begin{array}{ll}
u^k_i\geq \epsilon&\textrm{if }i\in\mathcal I_k\\
u^{k'}_i\leq -\epsilon&\textrm{if }i\in\mathcal I_k \textrm{ and }k'\neq k.
\end{array}
\right.
\end{equation}
This information is then used in an iterative PDE process with a time parameter $t$, in which we seek to minimise the  sum of normalised ratios $\sum_k \frac{\Delta_1 (u^k)}{|u^k|}$.  Denoting ${\bf u}= [u^1,\cdots u^L]$ and a time step $\Delta t>0$. Then formally, we seek to minimise:
\begin{equation}\label{pde}
{\bf u}^{(t+1)}
=\uargmin{{\bf u}} \frac{\| {\bf u}-{\bf u}^{(t)}\|^2}{2\Delta t}+\sum_{k=1}^L\left( \Delta_1(u^k)-\frac{\Delta_1 (u^{k,(t)})}{|u^{k,(t)}|}\langle \sign(u^{k,(t)}), u^k\rangle \right),\\
\end{equation}
under the set of   previously described coupling  and data \eqref{constr_data} constraints.
Following \cite{hein2013total,Feld2019}, a final shifting ${u}^{k,(t+1)}={u}^{k,(t+1)}-\textrm{median}({u}^{k,(t+1)})$ and  a  normalisation ${\bf u}^{(t+1)}={\bf u}^{(t+1)}/||{\bf u}^{(t+1)}||$ are necessary at the end of each iteration to prevent from converging to trivial solutions.

When a unique $u^k$ is considered, the scheme iteratively decreases the ratio $\frac{\Delta_1 (u^{k,(t)})}{|u^{k,(t)}|}$ since $\langle \sign(u^{k,(t)}), u^{k,(t)}\rangle=|u^{k,(t)}|$, so that the solution $u^{k,(t+1)}$ of \eqref{pde} necessarily satisfies:
\begin{equation}
\Delta_1 (u^{k,(t+1)})\leq \frac{\Delta_1 (u^{k,(t)})}{|u^{k,(t)}|}\langle \sign(u^{k,(t)}), u^{k,(t+1)}\rangle \leq \frac{\Delta_1 (u^{k,(t)})}{|u^{k,(t)}|}|u^{k,(t+1)}|.
\end{equation}
As noticed in \cite{Feld2019}, the  scheme  makes $u^{k,(t)}$ converge to a bivalued function that naturally segment the graph. As $L$  variables are coupled, the final labelling of a node $i$ is chosen from   the variable $u_i^k$  with the highest value: $y_i=\uargmax{k} u_i ^k$.

\textbf{Optimisation Scheme.} For each time step $t$, the problem \eqref{pde} is solved at successive time steps using the accelerated primal dual algorithm of \cite{chambolle2011first}. Denoting as ${\bf v}={\bf u}^{(t)}$ the current estimation and initialising ${\bf u}_0=\tilde{\bf u}_0={\bf v}$, $z^{k}_0=WD{-1}u^k_0$, the algorithm to obtain ${\bf u}^{(t+1)}$ with an iterative sequence ${\bf u}_{\ell}$ indexed by $\ell$ reads:
$$\left\{\begin{array}{lll}
z^{k}_{\ell+1}&=z^k_\ell+\sigma_\ell WD^{-1} {\tilde u}^k_{\ell}\\
z^{k}_{\ell+1}&=\frac{z^{k}_{\ell+1}}{\max(1,|z^{k}_{\ell+1}|)}\\
u^{k}_{\ell+1}&=\frac{u^k_{\ell}+\tau_\ell\Delta t\left(\frac{\Delta_1 (v^{k})}{|v^{k}|} \sign(v^{k})+D^{-1}W z^{k}_{\ell+1}\right)}{1+\tau_\ell\Delta t}\\
u^{k}_{\ell+1}&=\textrm{Proj}_C(u^{k}_{\ell+1})\\
\gamma_\ell&=1/\sqrt{1+\tau_\ell/\Delta t},\, \tau_{\ell+1}=\tau_\ell\gamma_\ell,\, \sigma_{\ell+1}=\sigma_\ell/\gamma_\ell\\
\tilde u^{k}_{\ell+1}&=u^{k}+\gamma_\ell(\tilde u^{k+1}- u^{k}),
\end{array}\right.$$
where the projection onto the set of constraints $C$ combining the coupled constraint and \eqref{constr_data} reads pointwise: \begin{equation}\label{proj_constr}
\textrm{Proj}_C(u^k_i)=\left\{\begin{array}{ll}
\max(u^k_i, \epsilon)&\textrm{if }i\in\mathcal I_k\\
\min(u^k_i,- \epsilon)&\textrm{if }i\in\mathcal I_{k'} \textrm{ and }k'\neq k.\\
u^k_i-\frac1L\sum_{k'}u^{k'}_i&\textrm{if }i>l.
\end{array}
\right.
\end{equation}
For positive parameters $\sigma_0$ and $\tau_0$ satisfying $\sigma\tau<4$, such process makes ${\bf u}_{\ell}$ converges to ${\bf u}^{(t+1)}$,
the solution of \eqref{pde}.

\section{Experimental Results}
This section is devoted to describe in detail the set of experiments that we conducted to validate our GraphX$^{\textbf{\scriptsize NET}}$ approach. \medskip

\noindent
\textbf{Data Description.} We evaluate our approach using the ChestX-ray14~\cite{wang2017chestx} dataset, which is composed of $112,120$ frontal chest view X-ray with size of $1024\times 1024$. The dataset is composed of 14 classes (pathologies).  All measurements were taken from this dataset.

\medskip
\noindent
\textbf{Evaluation Methodology.} We validate our theory as follows. Firstly, we visualise the graphical construction and classification tasks of our graph-based semi-supervised framework. Secondly, the main part of the evaluation is to compare our GraphX$^{\textbf{\scriptsize NET}}$  to the state-of-the-art  methods on X-ray classification. We compare ours against two deep learning techniques: WANG17~\cite{wang2017chestx} and YAO18~\cite{yao2018weakly}. To evaluate the classifier output quality of the compared approaches, we performed a ROC analysis using the area under the curve (AUC) per pathology along with their average. Finally, beside the official split, we perform a comparison with random partitions on ChestX-ray8 using WANG17~\cite{wang2017chestx} as baseline.

\begin{figure}[t!]
\centering
\includegraphics[width=1\textwidth]{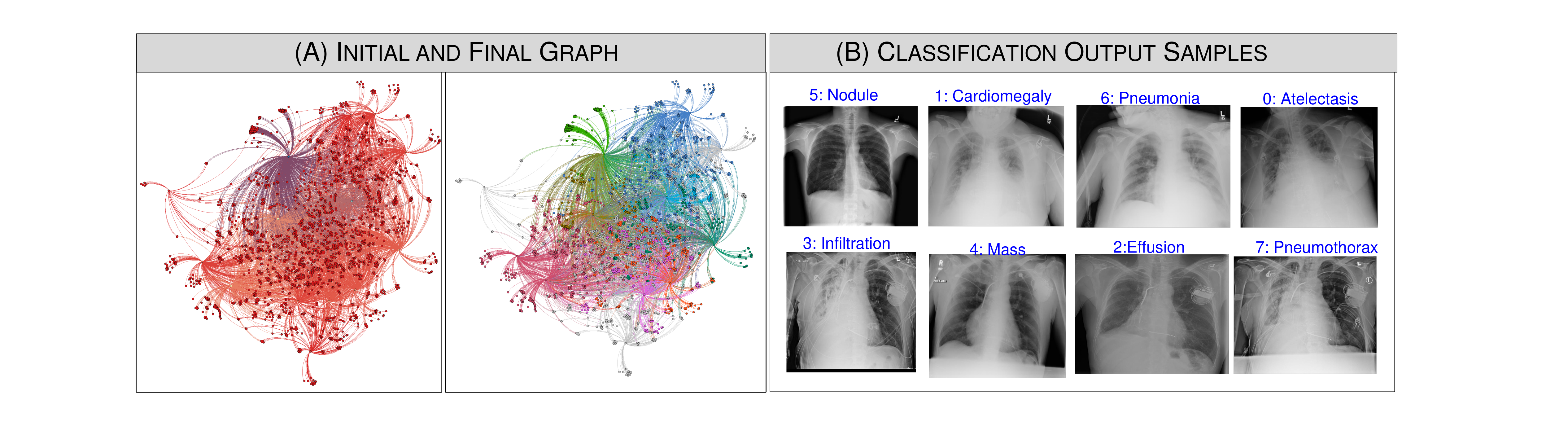}
\caption{Graphical Construction and Classification: (A) shows the graphical representation of the ChestX-ray14 dataset, where in the final classified graph, each colour represents a different class and (B) demonstrates examples of correct classifications produced by our framework.}
\label{fig::res2}
\end{figure}

\medskip
\noindent
\textbf{Results and Discussion.} Firstly, we start by giving some insight into our approach with some visualisations shown in Fig.~\ref{fig::res2}. The left side of the figure shows two graphs in which the first one illustrates the initial state of the graph created after computing the feature distances between the given X-ray data while the second one shows the graph after computing \eqref{pde}. The colours on the graph indicates an images belonging to a particular class. The right side shows few sample graph label output, that were correctly classified, of our approach.

\begin{table}[t!]
	\begin{minipage}{0.32\linewidth}
		\centering
		\includegraphics[width=1\textwidth]{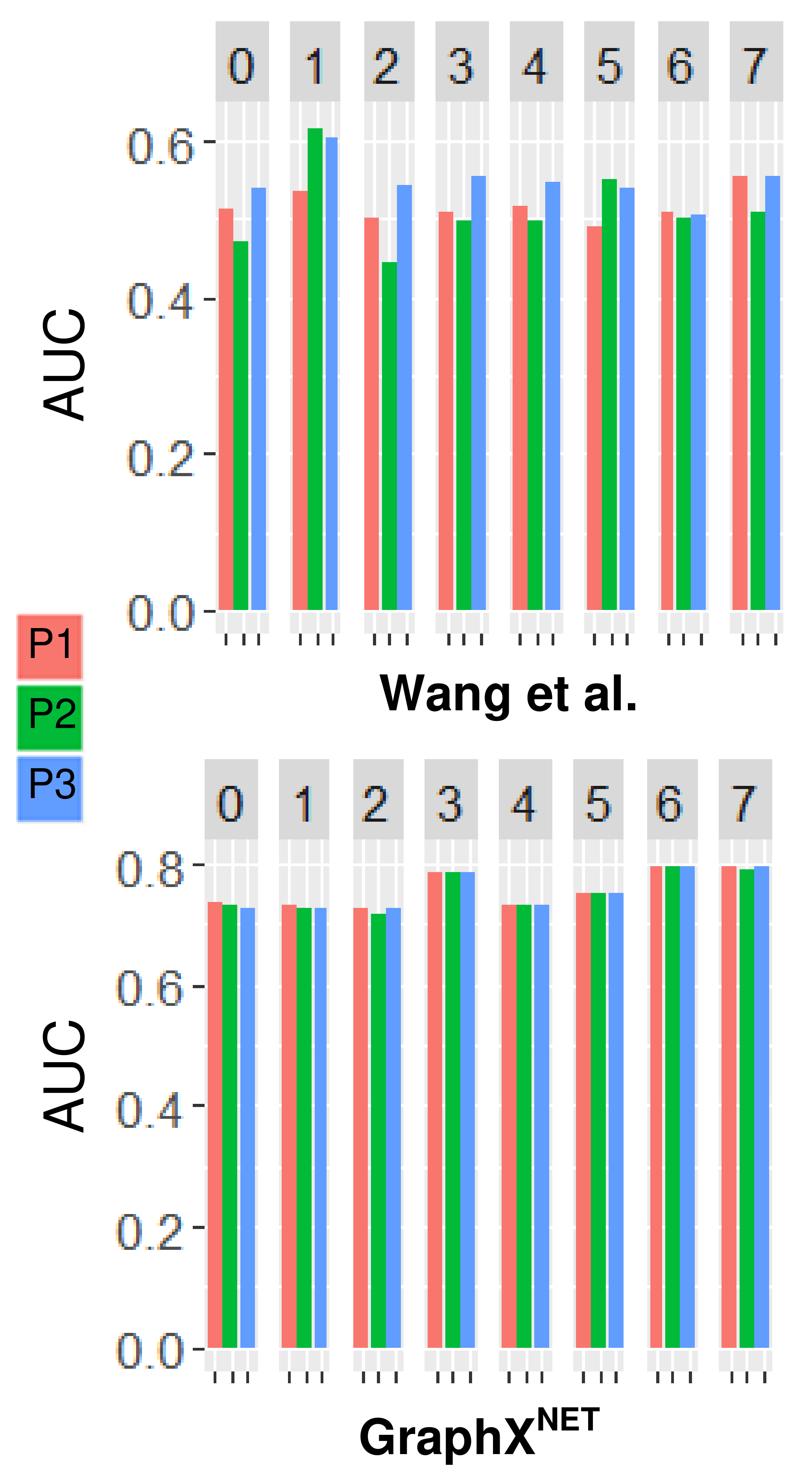}
		\label{ }
	\end{minipage}
		\begin{minipage}{0.6\linewidth}
		\captionof{table}{ Comparison of the classification accuracy of GraphX$^{\textbf{\scriptsize NET}}$ against two state-of-the-art deep learning method, Wang et al.~\cite{wang2017chestx} and Yao et al.~\cite{yao2018weakly}. Here we report the AUC measure over all 14 pathology classes along with the overall average. Plots on the left side highlight the sensitivity of the AUC for each class when changing the data partition of the data set (using $15\%$ for training)}
		\label{table:compre1}
		\centering
    \begin{tabular}{|l|c|c|c|}
    \hline
    \rowcolor[HTML]{EFEFEF}
    \multicolumn{1}{|c|}{\cellcolor[HTML]{EFEFEF}\textsc{Pathology}} & \textsc{Wang17}~\cite{wang2017chestx} & \textsc{Yao18}\cite{yao2018weakly} & GraphX$^{\textbf{\scriptsize NET}}$ \\ \hline
    Cardiomegaly & 0.81 & 0.856 & \cellcolor[HTML]{9AFF99}0.8799 \\ \hline
    Emphysema & 0.833 & \cellcolor[HTML]{9AFF99}0.842 & 0.8407 \\ \hline
    Edema & 0.805 & \cellcolor[HTML]{9AFF99}0.806 & 0.802 \\ \hline
    Hernia & 0.872 & 0.775 & \cellcolor[HTML]{9AFF99}0.8722 \\ \hline
    Pneumothorax & 0.799 & 0.805 & \cellcolor[HTML]{9AFF99}0.837 \\ \hline
    Effusion & 0.759 & \cellcolor[HTML]{9AFF99}0.806 & 0.792 \\ \hline
    Mass & 0.693 & 0.777 & \cellcolor[HTML]{9AFF99}0.809 \\ \hline
    Fibrosis & 0.786 & 0.743 & \cellcolor[HTML]{9AFF99}0.8034 \\ \hline
    Atelectasis & 0.7 & \cellcolor[HTML]{9AFF99}0.733 & 0.7189 \\ \hline
    Consolidation & 0.703 & 0.711 & \cellcolor[HTML]{9AFF99}0.7336 \\ \hline
    Pleural Thicken & 0.684 & 0.724 & \cellcolor[HTML]{9AFF99}0.757 \\ \hline
    Infiltration & 0.661 & 0.673 & \cellcolor[HTML]{9AFF99}0.7205 \\ \hline
    Nodule & 0.669 & \cellcolor[HTML]{9AFF99}0.718 & 0.7113 \\ \hline
    Pneumonia & 0.658 & 0.684 & \cellcolor[HTML]{9AFF99}0.7664 \\ \hline\hline
    \multicolumn{1}{|c|}{\textsc{Average} AUC} & 0.7451 & 0.7610 & \cellcolor[HTML]{9AFF99}0.7888 \\ \hline
    \end{tabular}
    	\end{minipage}
\end{table}

To evaluate the performance of our approach, we compared it against state of the art Deep Learning approaches, namely WANG17~\cite{wang2017chestx} and YAO18~\cite{yao2018weakly}. To the best of our knowledge, there are no semi-supervised learning method, for X-ray classification, that we can compare against. Therefore, we set as our baseline WANG17 and YAO18.
Table~\ref{table:compre1} shows the AUC results of the compared approaches where overall our approach outperformed the other methods across most pathology. Even though YAO18 performs better in some classes, a clear advantage of our approach over these two baselines is that while their approach rely in a huge percentage of data, 70\%, we were able to report a better average AUC result with only 20\% of the data.

Moreover, due to the semi-supervised nature of the GraphX$^{\textbf{\scriptsize NET}}$ framework, the classification output is very stable with respect to changes in the partition of the dataset. In the plots next to Table~\ref{table:compre1}, we tested the AUC of both the GraphX$^{\textbf{\scriptsize NET}}$ framework and WANG17 \cite{wang2017chestx} using three different random data partitions, including the partition suggested by Wang. The Wang method is very sensitive to changes in the partition due to the face that supervised methods are heavily reliant on the training set being representative. However, there is minimal change in the performance of GraphX$^{\textbf{\scriptsize NET}}$ over the three different partitions as the underlying graphical representation is invariant to the partition.

\begin{table}[t!]
	\begin{minipage}{0.59\linewidth}
		\centering
		\includegraphics[width=1\textwidth]{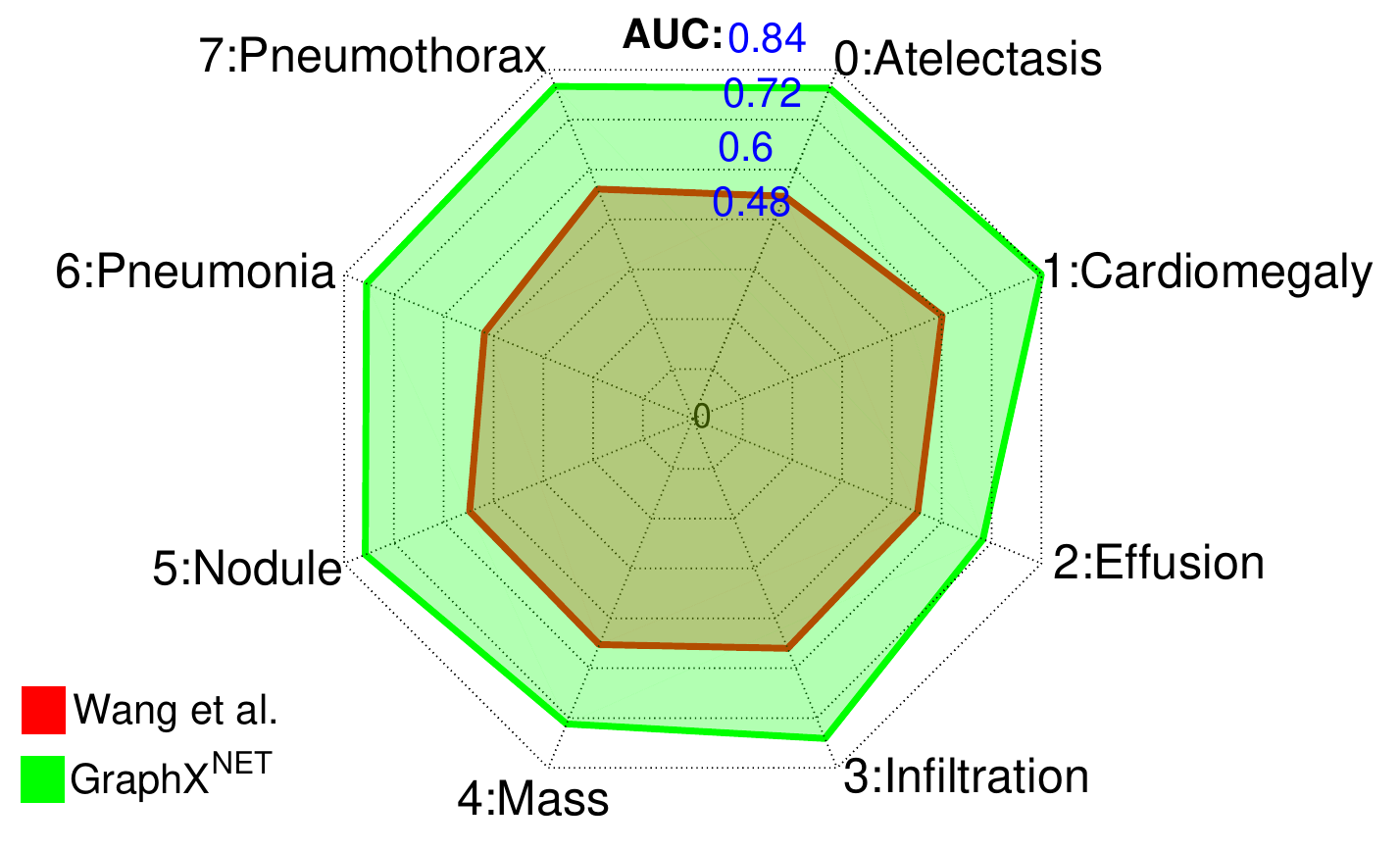}
		\label{}
	\end{minipage}
		\begin{minipage}{0.40\linewidth}
		\captionof{table}{Comparison of the classification accuracy of GraphX$^{\textbf{\scriptsize NET}}$ against a state-of-the-art deep learning method by Wang et al. \cite{wang2017chestx}. We give the average AUC measure over all eight classes using different amounts of labelled data. Additionally, we give a class by class comparison between the two methods using $70\%$  of the labelled data for the Wang method and $20\%$ for GraphX$^{\textbf{\scriptsize NET}}$.}
		\label{table:student}
		\centering
		\begin{tabular}{|c|c|c|c|c|c|}
            \hline
            \multicolumn{6}{|c|}{\cellcolor[HTML]{EFEFEF}\textsc{Wang17}~\cite{wang2017chestx}} \\ \hline
            \%Labeled             & \multicolumn{5}{c|}{70\%}                  \\ \hline
            AUC           & \multicolumn{5}{c|}{0.548}                \\ \hline \hline
            \multicolumn{6}{|c|}{\cellcolor[HTML]{EFEFEF}GraphX$^{\textbf{\scriptsize NET}}$}   \\ \hline
            \%Labeled              & 2\%   &  5\%      & 10\%   &  15\%     &   20\%     \\ \hline
            AUC     & 0.53     & 0.58     & 0.63     & 0.68    & 0.78    \\ \hline
        \end{tabular}
	\end{minipage}
\end{table}

For more detailed analysis of this dependency on the portioning and to further support the advantage of our GraphX$^{\textbf{\scriptsize NET}}$, in Table~\ref{table:student}, we compare the AUC produced by GraphX$^{\textbf{\scriptsize NET}}$ against WANG17 using a random split over ChestX-ray8. We find that GraphX$^{\textbf{\scriptsize NET}}$ produces a more accurate classification using $5\%$ of the data labels than the WANG17 method does using $70\%$ of the data labels. Furthermore, as we feed GraphX$^{\textbf{\scriptsize NET}}$ more of the data labels, the classification accuracy increases and becomes competitive against other the deep learning framework of that YAO18~\cite{yao2018weakly} whilst using a far smaller amount of data labels.

\section{Conclusion}

In this work, we tackled the problem of X-ray classification and introduced a novel semi-supervised framework based on a graph-based optimisation model, which is the first method that exploits graph-based semi-supervised learning for X-ray data classification. We also introduced a new multi-class classification functional with carefully selected class priors that allows for a smooth solution. 
We demonstrated that our method produces highly competitive results on the ChestX-ray14 data set whilst drastically reducing the need for annotated data.

\section*{Acknowledgments}
AIAI is supported by the CMIH, University of Cambridge. NP is supported by the European Union’s Horizon 2020 research and innovation programme under the Marie Skłodowska-Curie grant No 777826.
CBS acknowledges  Leverhulme Trust (Breaking the non-convexity barrier), the Philip Leverhulme Prize, the EPSRC grants  EP/M00483X/1 and EP/N014588/1, the European Union Horizon 2020, the Marie Skodowska-Curie grant 777826 NoMADS and 691070 CHiPS, the CCIMI and the Alan Turing Institute.

%
%
\bibliographystyle{splncs04}
\bibliography{bibMICCAI19}

\end{document}